\def\year{2021}\relax
\documentclass[letterpaper]{article} 
\usepackage{aaai21}  
\usepackage{times}  
\usepackage{helvet} 
\usepackage{courier}  
\usepackage[hyphens]{url}  
\usepackage{graphicx} 
\usepackage{multirow}
\usepackage{amsmath}
\usepackage{natbib}  
\usepackage{caption} 
\title{A Knowledge-Grounded Dialog System Based on \\ Pre-Trained Language Models}
\author{
    Weijie Zhang,
    Jiaoxuan Chen,
    Haipang Wu,
    Sanhui Wan,
    Gongfeng Li
    \\
}
\affiliations{
    Hithink RoyalFlush Information Network Co.,Ltd., Zhejiang, China \\
    \{zhangweijie2, chenjiaoxuan, wuhaipang, wansanhui, ligongfeng\}@myhexin.com
}

\begin{document}

\maketitle

\begin{abstract}
We present a knowledge-grounded dialog system developed for the ninth Dialog System Technology Challenge (DSTC9) Track 1 - \textit{Beyond Domain APIs: Task-oriented Conversational Modeling with Unstructured Knowledge Access}. We leverage transfer learning with existing language models to accomplish the tasks in this challenge track. Specifically, we divided the task into four sub-tasks and fine-tuned several Transformer models on each of the sub-tasks. We made additional changes that yielded gains in both performance and efficiency, including the combination of the model with traditional entity-matching techniques, and the addition of a pointer network to the output layer of the language model.
\end{abstract}

\section{Introduction}
Task-oriented dialog systems are developed to enable automated agents to help user complete specific tasks, such as booking movie tickets, hotels, restaurants, etc, and answer user’s questions with relevant information. In previous task-oriented dialog dataset, such as MultiWOZ 2.1 \cite{budzianowski2018multiwoz,eric2019multiwoz}, domain and entity information are typically stored in a structured database and can be retrieved via an API. The ninth Dialog System Technology Challenge (DSTC9) Track 1 \cite{kim2020domain} proposed another scenario wherein the user seeks domain or entity information which is not found in the structured database, but may exist in unstructured texts such as the description, FAQ and user review sections of the relevant web pages.

This challenge track introduced an augmented dataset based on MultiWOZ 2.1, with additional turns that require answers grounded on external unstructured knowledge. The track also proposed a baseline system, which follows a pipeline consisting of three sub-tasks: 1) knowledge-seeking turn detection, 2) knowledge selection, and 3) knowledge-grounded response generation.

In this paper, we explore a knowledge-grounded dialog system that makes use of state-of-the-art neural language models and surpasses the baseline performance. We further improved the model’s performance by employing traditional text-matching methods based on N-gram and edit distance so as to narrow down the matching candidates for the model, as well as by adding a pointer network \cite{vinyals2015pointer} on top of the model to improve response generation.

\section{Task Description}
The DSTC9 Track 1 aims at building a task-oriented dialog system that can answer questions using unstructured knowledge data, typically in the form of FAQ texts from relevant websites.  The full details of this task are described in the task proposal \cite{kim2020domain}. Here we briefly describe the three proposed sub-tasks, and mention the important statistics of the dataset used to develop and evaluate the system.

\subsection{Sub-task 1: Knowledge-seeking Turn Detection}
Compared to the original MultiWOZ 2.1 dataset, this augmented version has added new query turns into the conversations, the answers to which cannot be found in databases or APIs but exist in the unstructured form of FAQ texts (i.e. question-answer pairs). The first sub-task is to detect such query turns, in order for the system to answer such queries in the following sub-tasks. This sub-task is formulated as a binary classification problem - given the dialog history, the API/DB entries and the external knowledge sources, the model need to output whether the current query need access to external knowledge.

\subsection{Sub-task 2: Knowledge Selection}
For queries that require access to external knowledge, The model need to select the correct FAQ from all external knowledge sources. The selected FAQ is correct only if the FAQ is from the correct domain and the correct entity and the content of the FAQ matches the user’s question. This sub-task is formulated as a problem of text retrieval. The model need to output the ranked top-k FAQs.

\subsection{Sub-task 3: Knowledge-grounded Response Generation}
This sub-task is the final step in the pipeline - produce an answer to the current query, utilizing the external knowledge selected in the sub-task 2. The sub-task fits in the framework of text-to-text generation. High-quality answers should not only contain the correct information, but also appear natural and appropriate in the dialog context.

\subsection{Dataset}
The dataset used to train and evaluate our system encompasses conversations that provide tourist information about hotels, restaurants, taxi, etc. The training, validation and test data have 71348, 9663 and 4181 user turns, respectively, 19184 (26.9\%), 2673 (27.7\%) and 1981 (47.4\%) out of which are knowledge-seeking turns.

The knowledge sources for both the training and validation data include 4 domains (i.e. hotel, restaurant, taxi and train), 145 entities and 2900 FAQs. All domains and entities appearing in the validation data are seen in the training data. In contrast, the knowledge sources for the test data include 5 domains (added attraction domain), 668 entities and 12039 FAQs. Out of the 1981 knowledge-seeking turns in the test data, 1004 (50.7\%) are associated with domains or entities unseen in the training data.

\section{Related Work}
Since first proposed in 2017 \cite{vaswani2017attention}, Transformer models have achieved great success in various natural language processing (NLP) tasks in recent years. This type of neural network architecture, unlike the popular predecessors - recurrent neural networks (RNNs) and convolutional neural networks (CNNs), relies on the ``attention" mechanism to calculate the interaction between inputs across the time dimension, which contributes to its better handling of long-distance dependency and alleviates the problem of vanishing gradient often encountered by RNNs \cite{hochreiter2001gradient}.

Transfer learning, on the other hand, further pushes the model capacity by pre-training the model on large amount of unlabeled texts before fine-tuning on various NLP tasks. This has been shown repeatedly to increase the performance and robustness of the model on downstream tasks. The GPT-2 model \cite{radford2019language}, consists of one Transformer stack as decoder and was pre-trained with the language model objective, which is to predict the next token in an autoregressive manner. BERT \cite{devlin2018bert}, in contrast, consists one single encoder stack, and was pre-trained with the masked language model objective, which is to reconstruct the corrupted/masked tokens in the input. ALBERT \cite{lan2019albert} reduces the number of parameters in BERT by means of parameter sharing and matrix factorization. This helps regularize the model and was shown to have superior performance than BERT. ELECTRA \cite{clark2020electra} achieves higher pre-training efficiency (better performance with less pre-training steps), by first training a generator model to predict masked tokens in the input, and then training a discriminator model to predict whether each token is from the original input or the generator. Finally, T5 \cite{raffel2019exploring} is an encoder-decoder model pre-trained with span-corruption objective on a larger corpus, both of which contributed to better performance in experiments. When fine-tuned under the unified text-to-text framework, T5 was shown to achieve state-of-the-art performance in a wide range of NLP tasks.

The DSTC9 Track 1 proposed a baseline system based on BERT and GPT-2 \cite{kim2020domain}. Compared to traditional methods, including unsupervised anomaly detection, information retrieval methods and text extraction by rule, neural network models performed better on all sub-tasks in this challenge track, demonstrating the efficacy of transfer learning with Transformer models.

Nevertheless, some notable drawbacks of deep neural classifiers include high computational cost, as well as being over-confident over samples far away from training data \cite{hein2019relu}. Moreover, class imbalance in training data can result in degenerate models which perform poorly on sparse and difficult samples. One simple yet effective method to mitigate class imbalance is replacing cross-entropy loss with focal loss \cite{lin2017focal}, which down-weights the loss from easy samples and places more focus on difficult misclassified samples.

\section{Methodology}

\begin{figure*}[htb]
\centering
\includegraphics[width=0.95\textwidth]{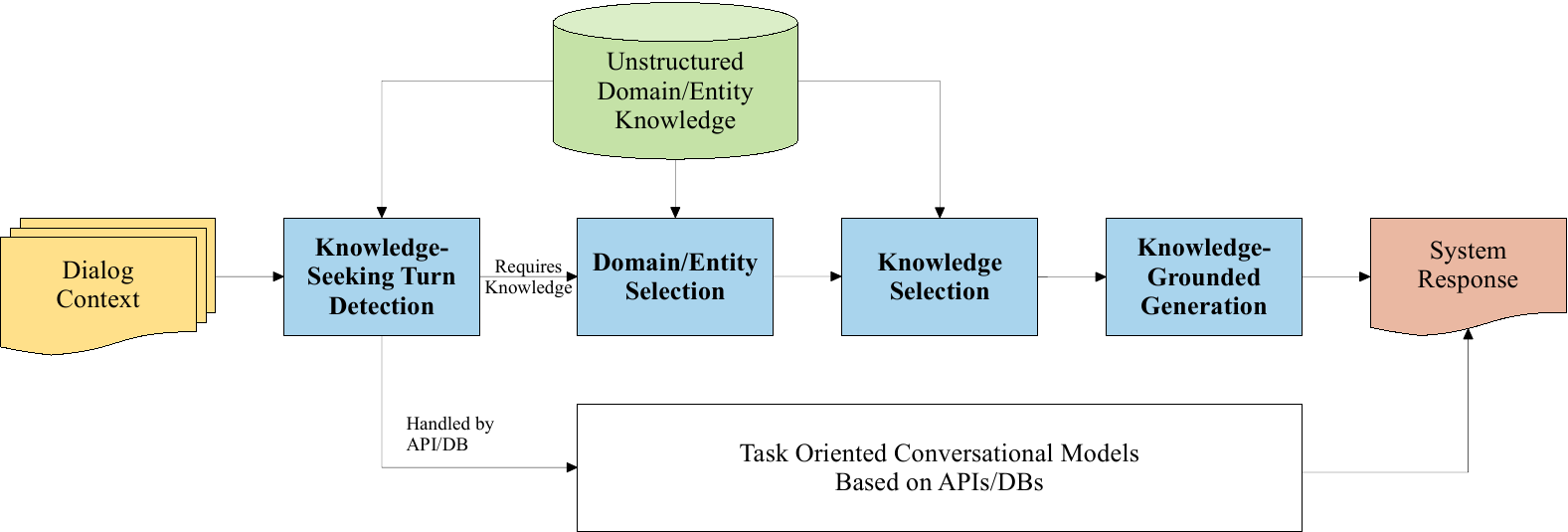}
\caption{Architecture of our knowledge-grounded dialog system. It is similar to the baseline except that we decouple domain/entity selection from knowledge selection. Note that we focus on the blue components in this diagram, and the conversational models based on APIs/DBs are out of the scope of this paper.}
\label{fig:system}
\end{figure*}

We followed the pipelined architecture of the baseline system and made tweaks in several areas. Most notably, we further split the sub-task of knowledge selection into two smaller sub-tasks: 1) domain/entity selection and 2) domain/entity-specific knowledge selection. Thus, our system consists of four sub-tasks, illustrated in Figure~\ref{fig:system}. Besides, in each of the sub-tasks, we replace BERT or GPT-2 used in the baseline system with newer Transformer models, such as ELECTRA and T5. Our approach in each sub-task is outlined in detail below.

\subsection{Sub-task 1: Knowledge-seeking Turn Detection}
We adopted the same approach as the baseline, training a neural network model as a binary classifier. In contrast with the baseline, we replace BERT with ELECTRA and T5, and cross-entropy loss with focal loss. For every user utterance, we use a Transformer to encode the text into an embedding vector. Then a linear layer projects the embedding into the class score. 

The focal loss serves as a drop-in replacement for the binary cross-entropy loss, in the hope of making the model more robust on difficult samples, and especially on the unseen test data. Even though the DSTC9 Track 1 dataset does not have heavy class imbalance as in the case of object detection, for which the focal loss was initially proposed \cite{lin2017focal}, some examples are inherently easier to learn than others. By down-weighting the loss on easy samples, there is less chance of over-fitting and making over-confident predictions on unseen data.  We use the non-$\alpha$-balanced variant of the focal loss with the hyperparameter $\gamma = 2$,
\begin{equation}
\text{FL} = -\sum_{i=1}^{n}t_i (1 - p_i)^\gamma \log(p_i)
\end{equation}
where $n$ is the number of classes, $t_i \in \{0, 1\}$ is the truth label and $p_i \in [0, 1]$ is the probability predicted by the model for the $i$-th class.

We only use the encoder stack of T5, even though it has an encoder-decoder architecture, and fine-tuned the encoder to predict the class labels directly, even though the original authors used the text-to-text approach on all tasks. This allows the T5 to be used in the same manner as other BERT-based models and has the advantage in the ease of implementation. Since T5 does not use the [\textit{CLS}] token in tokenization and pre-training, we take the encoded output corresponding to the [\textit{EOS}] token instead of the [\textit{CLS}] token.

\subsubsection{Decision Threshold}
After the model is trained, we look for a threshold value $t_\text{ktd} \in (0,1)$ for positive prediction. We optimize this value to be 0.45 on the validation data to maximize the F1 score, which is used in the official ranking criterion of DSTC9 Track 1. With this threshold, the recall in still lower than the precision on the validation data. Therefore, we also experiment a lower threshold value of 0.25 on the test data, to promote recall on unseen data.

\subsection{Sub-task 2: Knowledge Selection}
In early experiments, we tried the single-model approach in the baseline, where the user utterance is matched against every knowledge snippet under all domains and entities, and found it has several drawbacks. First, this approach is inhibitively slow in the inference phase, as the number of candidates equals the total number of knowledge snippets. Second, because there are similar FAQs under different domains and entities, for example many hotels has the FAQ ``Can I bring my dog?" the model tend to produce a high-confidence positive score, solely based on the similarity between the FAQ and the user utterance, ignoring the mismatch between the domain/entity and the dialog history.

Therefore, we opted to decouple domain/entity selection and knowledge selection, as illustrated in Figure~\ref{fig:system}. Two separate models are trained to carry out the two-stage selection - firstly all domains and entities are ranked according to the matching score against the dialog history, secondly only the knowledge snippets associated with the top-1 domain/entity are ranked according to the matching score against the current user utterance.

\subsubsection{Entity Matching}
For entity selection, we experimented statistical methods for entity matching, such as TF-IDF similarity, N-gram match and edit distance, to narrow down the candidates to be ranked by the deep model. Knowledge snippets associated with entities that do not appear in the dialog history will be filtered out. For knowledge snippets associated with domains (e.g. train, taxi) where there is no specific entities, we do not use the statistical methods and always add these snippets to the set of candidates.

In early experiments, we calculated TF-IDF similarity between the entity and the dialog history and kept only the top-10 ranked entities. This approach was subsequently superseded by a method based on N-grams and edit distance, which proved to offer higher precision and recall than TF-IDF in our experiments.

\textbf{N-gram match.} We first apply word-level tokenization and N-gram search to match entity names in the dialog history. To minimize the number of false positives, we set two threshold values $t_1$ and $t_2$, the maximum document frequencies (df) of the N-gram counted on all dialog utterances and all entity names, respectively. In other words, the N-gram must appear in no more than $t_1$ utterances and no more than $t_2$ entity names, in order to be eligible for matching. Formally, we define the set of N-grams used for the $j$-th entity $e_j$ as
\begin{equation}
W_j = \left\{ w_i^n | w_i^n \in e_j, \text{df}(w_i^n, U) < t_1, \text{df}(w_i^n, E) < t_2 \right\}
\end{equation}
where $w_i^n$ denotes an N-gram consisting of $n$ words, $E$ and $U$ denote the set of entity names and the set of all utterances, respectively. The set of entity candidates for a given dialog context $U_t$ can then be obtained as
\begin{equation} \label{eq:1}
E_t = \left\{ e_j| U_t \cap W_j \ne \emptyset  \right\} \text{.}
\end{equation}

We use $n \in \{1, 2, 3, 4\}$ and optimized the threshold values on the training data to be $ t_1 = N_\text{utt} / 100$ and $t_2 = 5$, respectively, where $N_\text{utt}$ is the total number of utterances which depends on the dataset being used (e.g. training, validation or test data).

\textbf{Fuzzy match with edit distance.} To account for typos in dialog texts, we allow N-grams that appear in the dialogs but not in the entity names to be matched as long as they are within a certain edit distance $t_\text{d}$ with the entity N-grams $W_j$ and have an utterance frequency lower than $t_3$,
\begin{equation}
\begin{split}
\tilde{W_j} = \left\{ w_k^n | w_k^n \in U, \text{d}_{\min}^{j, k} \le t_\text{d}, \text{df}(w_k^n, U) < t_3 \right\}, \\
\text{d}_{\min}^{j, k} = \min_{w_i^n \in W_j} \text{d}(w_k^n, w_i^n)
\end{split}
\end{equation}
where we use $n \in \{1, 2\}$. For uni-grams, we allow a maximum edit distance of 2, and for bi-grams a maximum edit distance of 1, i.e.
\begin{equation}
t_\text{d} = 
\begin{cases}
2 & \text{if } n = 1 \\
1 & \text{if } n = 2
\end{cases}
\text{.}
\end{equation}
The threshold $t_3$ was optimized on the training data to be $N_\text{utt} / 2000$. Note that this is much lower than $t_1$, because words with typos rarely re-occur with the same spelling. The set of N-grams $\tilde{W_j}$ can then be combined with $W_j$ in Equation~\ref{eq:1} to improve recall.

\subsubsection{Ranker Model}
We train a point-wise ranker using the same binary classifier model as in sub-task 1, by fine-tuning a range of Transformer language models including ALBERT, ELECTRA and T5, and compare their performance to the BERT-based baseline. Similar to sub-task 1, we fine-tune the T5 encoder only, and use the focal loss to replace the binary cross-entropy loss. As the input to the model, the dialog history is truncated to the last 256 tokens and concatenated with the domain and entity names for domain/entity selection, or both the question and the answer of the FAQ for knowledge selection. For the latter, the dialog history is further truncated to the last utterance only.

We use a negative sampling ratio of 1:1. For domain/entity selection, we randomly select a negative sample from the filtered candidates. For knowledge selection, we randomly choose one negative sample from the top-10 negative samples according to their TF-IDF similarity with the positive sample.

\subsubsection{Knowledge  Embedding}
Beside the approach described above, wherein the knowledge and the dialog history are concatenated together and then encoded by the model, we also tried encoding them separately and then calculating the cosine similarity between the embedded vectors. This approach has the benefit of less computation and faster speed in the test/inference phase, as all knowledge embeddings can be pre-computed at once.

We fine-tune BERT as the encoder with the following loss similar to the triplet loss \cite{weinberger2009distance,schroff2015facenet}, given a context embedding $x_i^a$ (anchor), the correct knowledge embedding $x_i^p$ (positive) and a false knowledge embedding $x_i^n$ (negative),
\begin{equation}
\begin{split}
& L = \sum_{i}^{N}\left[ \text{sim}(x_i^a, x_i^n) - \text{sim}(x_i^a, x_i^p)+ \alpha \right]_+, \\
& \text{sim}(A, B) = \cos(\theta) = \frac{A\cdot B}{\left\| A \right\|\left\| B \right\|} \text{.}
\end{split}
\end{equation}
We set the margin $\alpha$ to be 0.2 in our experiments. The negative sample is selected by semi-hard negative mining \cite{schroff2015facenet},
\begin{equation}
\beta < \text{sim}(x_i^a, x_i^p) - \text{sim}(x_i^a, x_i^n) < \alpha
\end{equation}
where $\beta$ is set to 0.01. This ensures the negative sample lies close to the positive sample within the margin $\alpha$, which can be helpful for training efficiency.

At test time, we rank all knowledge snippets according to the cosine similarity scores between the knowledge embeddings and the context embedding.

\subsection{Sub-task 3: Knowledge-grounded Response Generation}
We experimented T5, as its encoder-decoder architecture and text-to-text objective makes it naturally suited for this text generation task. As the input to the model, we concatenate the last two utterances in the dialog history, the top-1 ranked knowledge snippet, and the corresponding domain and entity name. The model is fine-tuned with language model objective and we use greedy decoding in the test phase.

\subsubsection{T5 pointer-generator network}
Due to the nature of the task, in order to answer the user’s question in an accurate and informative manner, the response usually contain parts of the knowledge snippet and the corresponding entity name, both of which are given as inputs to the model. Therefore, we try adding a pointer network \cite{vinyals2015pointer} on top of T5 to encourage the model to copy words directly from the input, rather than selecting from a vast number of words from the vocabulary.

Similar to \cite{see2017get}, our pointer network first calculates the attention score $a^t$  of the current decoder output $s^t$ over all encoder outputs $h^t$, then the generation probability $p_{\text{gen}} \in [0,1]$ is computed as,
\begin{equation}
\begin{split}
p_{\text{gen}} & = \sigma \left( w_{\text{ptr}}[h_t^*; s_t] + b_{\text{ptr}} \right), \\
h_t^* & = \sum_i a_i^th_i^t
\end{split}
\end{equation}
where $\sigma$ is the sigmoid function, $w_{\text{ptr}}$ and $b_{\text{ptr}}$ are learned parameters and $h_t^*$ is the context vector. The final probability distribution of the generated token is computed as the weighted sum of the attention distribution and the vocabulary distribution,
\begin{equation}
P(w) = p_{\text{gen}}P_{\text{vocab}}(w) + (1 - p_{\text{gen}})\sum_{i:w_i=w}a_i^t \text{.}
\end{equation}

\subsection{Experimental set-up}
Our training and evaluation pipeline uses the baseline code \cite{kim2020domain}, with modifications needed to carry out our experiments. As with the baseline, we use the \textit{transformers} library \cite{wolf2019huggingface} to obtain the pre-trained models. We use the \textit{base} size for all models, which uses the 12-layer Transformer stack with hidden dimensionality of 768. On every sub-task, the models are fine-tuned for 10 epochs with a batch size of 32. The Adam optimizer \cite{kingma2014adam} is used with a constant learning rate of 1e-4 and a linear warm-up schedule during the first 500 iterations. A model checkpoint is saved after each epoch, and the best checkpoint is picked based on the validation results.

\subsection{Model Ensemble}
Bagging \cite{breiman1996bagging} ensemble is a common practice to boost the performance of individual models. For sub-task 1 and 2, we trained 5 models on different random seeds and aggregate individual predictions by voting in the test phase.

\section{Results and Discussion}
The overview of DSTC9 \cite{gunasekara2020overview} reports the performance of our system (Team 11) at various stages of the pipeline, as summarized in Table~\ref{tab:overview}. Here we also assess our methods in individual sub-tasks separately, decoupling the results on downstream  tasks from upstream tasks.

\begin{table*}[htb]
\centering
\begin{tabular}{l|c|c|ccc|cc}
           & Sub-task 1      & Sub-task 2      & \multicolumn{3}{c|}{Sub-task 3}                     & \multirow{2}{*}{Accuracy} & \multirow{2}{*}{Appropriateness} \\
           & F1              & R@1             & BLEU-1          & METEOR          & Rouge-L         &                           &                                  \\ \hline
Baseline   & 0.9455          & 0.6201          & 0.3031          & 0.2983          & 0.3039          & 3.7155          & 3.9386          \\
Our system & \textbf{0.9675} & \textbf{0.8702} & \textbf{0.3743} & \textbf{0.3854} & \textbf{0.3797} & \textbf{4.2722} & \textbf{4.2619}
\end{tabular}
\caption{Overall performance of our system (Team 11) compared to the baseline, as reported in the overview of DSTC9 \cite{gunasekara2020overview}. The accuracy and appropriateness of the system response were evaluated by human on a scale of 1-5. Note that the scores for the downstream sub-tasks are weighted by the performance of the upstream tasks.}
\label{tab:overview}
\end{table*}

\begin{table*}[htb]
\centering
\begin{tabular}{l|ccc|ccc}
                          & \multicolumn{3}{c}{Valid} & \multicolumn{3}{|c}{Test} \\
                          & Prec   & Recall  & F1     & Prec   & Recall  & F1    \\ \hline
Baseline (BERT)                  & 0.989  & 0.994   & 0.991  &        &         & 0.946 \\
T5, $t_\text{ktd} = 0.45$           & 0.996          & 0.994          & \textbf{0.995} & 0.990          & 0.919          & 0.953          \\
T5 ensemble, $t_\text{ktd} = 0.45$ & \textbf{0.997} & 0.994          & \textbf{0.995} & \textbf{0.993} & 0.934          & 0.963          \\
T5 ensemble, $t_\text{ktd} = 0.25$ & 0.992          & \textbf{0.995} & 0.993          & 0.988          & \textbf{0.948} & \textbf{0.968}
\end{tabular}
\caption{Validation and test results on sub-task 1: knowledge-seeking turn detection.}
\label{tab:task1}
\end{table*}

\begin{table}[htb]
\centering
\begin{tabular}{llccc}
                                              & \multicolumn{1}{c}{}      & \begin{tabular}[c]{@{}c@{}}Total \#\\ Entities\end{tabular} & \begin{tabular}[c]{@{}c@{}}Avg. \#\\ Matched\end{tabular} & Recall \\ \hline
\multirow{3}{*}{N-gram} & Train & 145              & 4.651              & 0.9978 \\
                                              & Valid & 145              & 4.420              & 0.9963 \\
                                              & Test & 668              & 7.562              & 0.9965 \\ \hline
\multirow{3}{*}{\begin{tabular}[c]{@{}l@{}}N-gram w/ \\ edit distance\end{tabular}}                 & Train & 145              & 4.887              & 0.9994 \\
                                              & Valid & 145              & 4.597              & 0.9978 \\
                                              & Test & 668              & 8.896              & 0.9980
\end{tabular}
\caption{Statistics of entity matching based on N-gram and edit distance, on training, validation and test data.}
\label{tab:task2_1}
\end{table}

\begin{table}[htb]
\centering
\begin{tabular}{lcc}
                   & R@1 - Valid    & R@1 - Test     \\ \hline
ALBERT             & 0.972          & 0.848          \\
ELECTRA            & 0.983          & 0.936          \\
T5                 & 0.985          & 0.962          \\
T5 (cross-entropy loss)        & 0.985          & 0.957          \\
T5 (all candidates) & 0.960          & 0.922          \\
T5 ensemble       & \textbf{0.986} & \textbf{0.970}
\end{tabular}
\caption{Validation and test results on sub-task 2 stage 1: entity selection. All models used only the candidates matched with N-gram and edit distance as inputs, unless stated otherwise (all candidates). All models were trained with focal loss, unless stated otherwise.}
\label{tab:task2_2}
\end{table}

\begin{table}[htb]
\centering
\begin{tabular}{lcc}
                   & R@1 - Valid    & R@1 - Test     \\ \hline
ALBERT             & 0.957          & 0.891          \\
ELECTRA            & \textbf{0.977} & 0.898          \\
T5                 & 0.970          & 0.913          \\
T5 (cross-entropy loss)      & 0.975          & 0.918          \\
T5 ensemble       & 0.976          & \textbf{0.923} \\
BERT (triplet loss) & 0.976          & 0.902         
\end{tabular}
\caption{Validation and test results on sub-task 2 stage 2: domain/entity-specific knowledge selection. All models were trained with focal loss, unless stated otherwise.}
\label{tab:task2_3}
\end{table}

\begin{table}[htb]
\centering
\begin{tabular}{lccc}
                                  & BLEU-4          & METEOR          & Rouge-L         \\ \hline
Baseline (GPT-2) & 0.1008          & 0.3884          & 0.2982          \\
GPT-2            & 0.1236          & 0.4134          & 0.3183          \\
T5                 & 0.1405          & 0.4383          & 0.3347          \\
T5-pointer     & \textbf{0.1459} & \textbf{0.4454} & \textbf{0.3432}
\end{tabular}
\caption{Automated evaluation results on validation data for sub-task 3: knowledge-grounded response generation. All models use greedy decoding except for the baseline.}
\label{tab:task3}
\end{table}

\subsection{Sub-task 1: Knowledge-seeking Turn Detection}

Both the baseline model and our T5 model achieved above 99\% F1 score on the validation data, shown in Table~\ref{tab:task1}, which is the reason why we did not explore other models and approaches. After the test data were released, however, it is evident that there is still room for improvement in terms of generalizability, marked by the relatively low recall on unseen data. Model ensemble was able to improve recall slightly, so was lowering the threshold value  $t_\text{ktd}$. Out of 24 participant teams, our system (\textit{T5 ensemble, $t_\text{ktd} = 0.25$}) was ranked 11th in this sub-task \cite{gunasekara2020overview}.

\subsection{Sub-task 2: Knowledge Selection}

The effectiveness of entity filtering based on N-gram and edit distance is demonstrated in Table~\ref{tab:task2_1}. It drastically reduced the number of candidates for the model, while retaining over 99.7\% of the correct answers. Table~\ref{tab:task2_1} also shows that N-gram match alone can recall over 99.6\% of the entities, while edit distance further improved the recall marginally by about 0.15\%. When combined with the model, statistical methods were shown to improve the accuracy by 2.5\% and 4.0\% on validation and test data (\textit{T5} vs. \textit{T5, all candidates} in Table~\ref{tab:task2_2}), respectively. 

In Table~\ref{tab:task2_2}, we show the performance of different models on the task of entity selection. ELECTRA and T5 are both very competitive on the validation data, while on the test data T5 outperformed the rest significantly. Model ensemble again boosted the performance compared to the single model.

Table~\ref{tab:task2_3} compares different models on knowledge selection. As seen previously, T5 was close to ELECTRA on the validation data, but was able to generalize better with the unseen test data. Knowledge embeddings (\textit{BERT, triplet loss}) performed only slightly worse than T5 on the test data, and remained an efficient solution if inference speed is of high priority.

Besides, the use of focal loss in training did not bring a clear benefit in this task. It led to slightly better performance in domain/entity selection (\textit{T5} vs. \textit{T5, cross-entropy loss} in Table~\ref{tab:task2_2}), but lowered performance in knowledge selection (Table~\ref{tab:task2_3}). The difference between focal loss and cross-entropy loss was relatively small (0.5\%).

On test data, our system (\textit{T5 ensemble}) was ranked 9th out of 24 teams in knowledge selection \cite{gunasekara2020overview}.

\subsection{Sub-task 3: Knowledge-grounded Response Generation}
In this particular task, greedy decoding outperformed nucleus sampling \cite{holtzman2019curious} used in the baseline system \cite{kim2020domain} according to automated evaluation metrics, as seen in Table~\ref{tab:task3} (\textit{Baseline} vs. \textit{GPT-2}). T5 performed better than GPT-2, while the T5-pointer model achieved the highest scores on validation data.

On test data, our system (\textit{T5-pointer}) was ranked 7th, 7th, and 2nd out of 24 teams in BLEU-4, METEOR and ROUGE-L scores, respectively \cite{gunasekara2020overview}, demonstrating the efficacy of the T5-pointer model.

\subsection{System Performance}
Our final system consists of only T5 models, fine-tuned on different sub-tasks. Table~\ref{tab:overview} shows the overall effectiveness of our system compared to the baseline system. Our system made significant improvements in all sub-tasks and consequently is able to produce response of higher quality, in terms of accuracy and appropriateness according to human evaluation.

Our experimental results show that the newer pre-trained language models were able to outperform the older models (BERT and GPT-2), thanks to recent advances including more efficient pre-training (ELECTRA), a more effective training objective and a larger pre-training corpus (T5). These advantages were carried into different downstream tasks. Meanwhile, in all sub-tasks of this challenge track, our system saw performance degradation on the test data compared to the validation set, indicating room for improvement in generalizability on unseen domains and locales.

\section{Conclusion}
We developed a knowledge-grounded dialog system for DSTC9 Track 1, which followed the pipelined architecture in the baseline, and outperformed the baseline in the overall performance. More importantly, we demonstrated that this level of performance can be achieved with relatively low computational cost. We employed several methods to improve the efficiency of training and inference. We tapped into the power of pre-trained language models which we fine-tuned for our task. Moreover, we decoupled the entity selection from knowledge selection, and employed statistical methods based on N-gram and edit distance to filter out irrelevant entities, thereby drastically reducing the number of candidates and hence the computational cost. Finally, our system was ranked 4th out of 24 participant teams in automated evaluation and 10th out of 12 finalist teams in human evaluation of the response quality \cite{gunasekara2020overview}.

\bibliography{references}

\end{document}